\documentclass{ifacconf}

\usepackage[utf8]{inputenc}
\usepackage{amsmath}

\usepackage{amsfonts}
\usepackage{amssymb}
\usepackage{graphicx}
\usepackage{natbib}

\usepackage{mathtools}
\usepackage{bm}
\usepackage{xcolor}

\newcommand{\fb}[1]{\textcolor{black}{#1}}
\begin{document}
\begin{frontmatter}

\title{Whole-Body Control of a Mobile Manipulator for Passive Collaborative Transportation}

\author[First]{F. Benzi}
\author[First]{C. Mancus}
\author[First]{C. Secchi}

\address[First]{Department of Sciences and Methods of Engineering,
    University of Modena and Reggio Emilia, Italy (e-mail: federico.benzi@unimore.it, cristian.mancus@unimore.it, cristian.secchi@unimore.it)}

% \setlength{\textfloatsep}{0.5\baselineskip plus 1.5\baselineskip minus 0.9\baselineskip}

% \maketitle

\begin{abstract}

    Human-robot collaborative tasks foresee interactions between humans and robots with various degrees of complexity. Specifically, for tasks which involve physical contact among the agents, challenges arise in the modelling and control of such interaction. In this paper we propose a control architecture capable of ensuring a flexible and robustly stable physical human-robot interaction, focusing on a collaborative transportation task. The architecture is deployed onto a mobile manipulator, modelled as a whole-body structure, which aids the operator during the transportation of an unwieldy load. Thanks to passivity techniques, the controller adapts its interaction parameters online while preserving robust stability for the overall system, thus experimentally validating the architecture.
    
\end{abstract}


\end{frontmatter}

\section{Introduction} \label{sec: intro}

The modern industrial paradigm foresees an increasingly close interaction among humans and robots. This is most strongly manifested in collaborative applications, in which the robot assists or cooperates with the human in order to accomplish a common goal.

During some collaborative tasks, a physical contact between the robot and its surroundings, these being the human or the environment, might take place. This contact can take place either directly onto the surface of the robot, or via the manipulation of a collaboratively held object. In this paper, we focus onto these types of contact, aiming at rendering such interactions both flexible and robust to different environments and parameter variations.

Different approaches in literature have been proposed for a safe and reliable collaborative load transportation. A vastly employed one is to teach the robot the collaborative task by means of programming by demonstration (PbD) \cite{calinon2009learning},\cite{gribovskaya2011motion}, in which the robot learns by encoding the human demonstrations as a task. The encoding process of the task model can be carried out via probabilistic frameworks based on Gaussian Mixture Model (GMM) \cite{calinon2007learning} or Hidden Markov Model \cite{hersch2008dynamical}. However, except for a certain degree of adaptation during the task execution, the performance of these controllers drop significantly whenever the interaction varies strongly from the demonstrated data, since their principle of operation is based on the local knowledge of the task. Thus, their generalization capabilities and consequently their flexibility are limited, making those non-ideal for dynamic environments. Additionally, learning-based methods also do not generally include the possibility of additional tasks, such as safety and joint limits control. Finally, processes as task model training and parameters tuning can result time consuming.

Other approaches directly integrate haptic information into an impedance or admittance controller \cite{agravante2013human, tagliabue2017collaborative} in order for the robot to comply with the external forces acting onto the manipulated object. Impedance and admittance control \cite{siciliano2012robot} are two commonly deployed strategies in interaction control. These controllers ensure compliance of the robot during the interaction phase by enforcing a dynamical behaviour in the form of a mechanical impedance, characterised by desired stiffness, damping and inertia parameters. In this way, the robot can adapt its behaviour according to haptic information and guarantee a stable interaction. In \cite{tagliabue2017collaborative} \fb{two Micro Aerial Vehicles (MAVs)} perform a collaborative transportation task based on the master-slave paradigm, in which the slave complies to the external force applied by the master to the payload via an admittance controller. The inherent passivity of the dynamics ensures a robustly stable interaction \cite{secchi2007control}.

Adapting the dynamic parameters during the interaction can be greatly beneficial for the application, since it allows for an higher flexibility of the behaviour, of peculiar interest for collaborative applications \cite{dimeas2016online}. This is carried out e.g. in \cite{gribovskaya2011motion}, in which an adaptive impedance is utilised to compensate for unmodeled uncertainties during the collaborative task execution due to the human behaviour. In here, however, there is no formal stability analysis of the adaptive impedance. In fact, varying online the dynamic parameters may lead to the loss of passivity \cite{ferraguti2015energy} and unstable behaviours might be implemented during the human-robot interaction \cite{ferraguti2019variable}.

The formulation of a provably stable variable admittance controller has been successfully carried out in \cite{secchi2019energy} by means of an optimization framework. The exploitation of energy tanks \cite{ferraguti2015energy, franken2011bilateral} allows to separate passivity from set dynamics, treating the energy flow in the tank as a requirement for the passivation of any desired behaviour.

The architecture was successively augmented in \cite{benzi2021optimization}, allowing for the simultaneous execution of multiple additional tasks alongside the passivisation of the behaviour. In here, the tasks are encoded by means of Control Barrier Functions (CBFs) \cite{ames2019control} following a procedure akin to \cite{notomista2020set}. CBFs have been deployed in robotics in order to constraint the system inside a subset of its state space, thus enforcing constraints onto the behaviour of the robot. Via time-varying CBFs \cite{notomista2020persistification} it is also possible to enforce time-varying constraint, which are well suited for safety applications. The control input satisfying the set of constraint can be found as the solution of a convex optimization problem, ensuring the real-time capabilities of the architecture. Nevertheless, only simple robotic arms have been considered so far.

Subsequent evolutions of the architecture are presented in \cite{benzi2022shared}, in which the control structure is deployed for passively implementing shared autonomy in a multi-robot teleoperation scenario, and in \cite{ferrari2022icra} for an effective and seamless human-robot collaboration via a bidirectional communication channel.

In this paper, we aim at deploying the architecture presented in \cite{benzi2021optimization}, further augmenting it for achieving the whole body control of the robot, in order to reproduce a collaborative human-robot transportation task in a robustly stable way, while performing additional tasks. This novel approach allows us to overcome the rigidity of standard PbD approaches, ensuring a flexible and adaptable behaviour at all times, while preserving the safety of all the parties involved.

The robot chosen for the application is a custom mobile manipulator, consisting of a differential drive mobile platform with a robotic arm mounted on top, modeled as a whole body system.
A similar task was implemented in \cite{nozaki2009motion}, but the arm and the mobile base were modelled separately and the impedance dynamics was fixed.

Thus, the contributions of this paper are:
\begin{itemize}
    \item A control architecture capable of performing a collaborative human-robot transportation task in a flexible and robust way
    \item A whole-body kinematic model and control of the robot, capable of considering non-holonomic constraints introduced by the mobile base
    \item The possibility of implementing multiple tasks alongside the main transportation one
\end{itemize}

The paper is organised as follows: in Sec. \ref{sec: prob} the main problem addressed in this paper is formulated. In Sec. \ref{sec: kin_model} the whole-body model of the mobile manipulator is laid out. In Sec. \ref{sec: tanks} the collaborative constraint-oriented control architecture is presented. In Sec. \ref{sec: experiments} the architecture is experimentally validated and finally in Sec. \ref{sec: conclusions} conclusions are drawn and future work is discussed. 
\section{Problem Formulation} \label{sec: prob}
\begin{figure}[tb]
    \centering
    \includegraphics[height=3cm ,width=0.9\columnwidth]{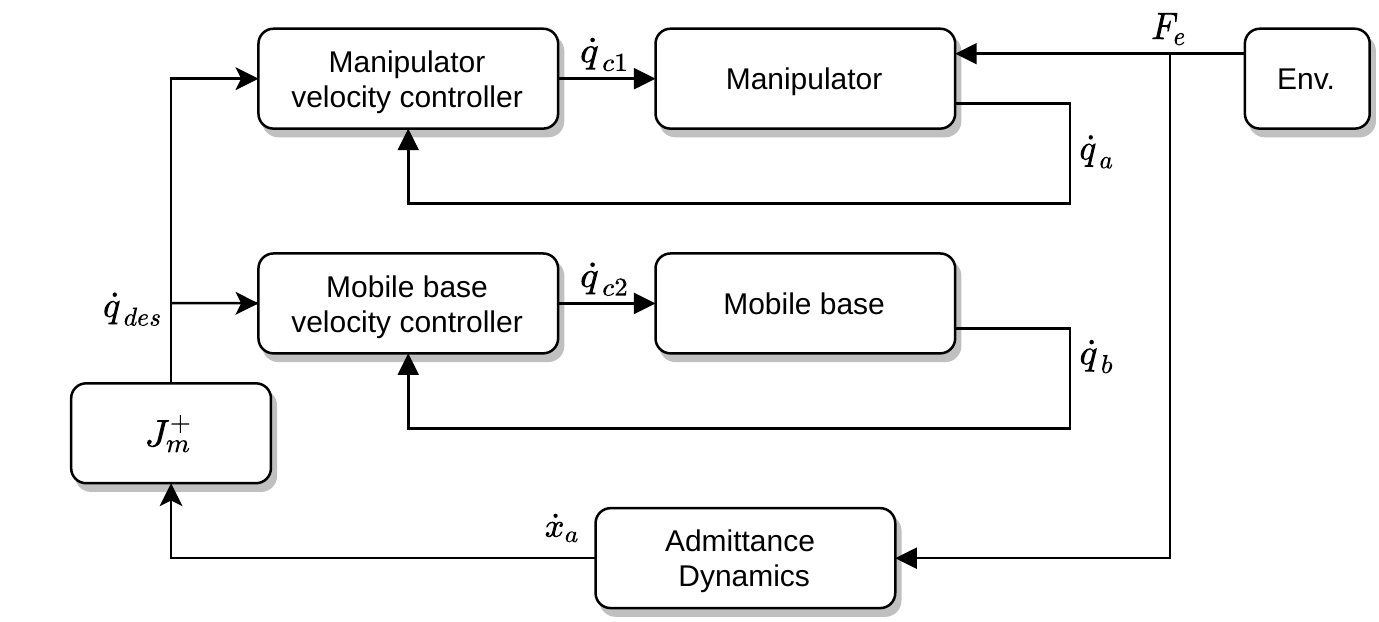}
    \caption{Whole-body admittance control architecture. We indicate as $\dot{\textbf{q}}_{c1}$ and $\dot{\textbf{q}}_{c2}$ the setpoint for the velocity controllers of the manipulator and the mobile base respectively.}
    \label{fig: adm_scheme}
\end{figure}

%\begin{itemize}
    %\item Kinematic model of manipulator and mobile base
    %\item Admittance dynamics
    %\item Time-varying admittance dynamics
    %\item Additional Tasks(?) \fb{\textbf{this requires further work and experiments}}
    %\item We aim at...
%\end{itemize}
Let us consider a velocity controlled mobile manipulator composed of a fully actuated $n_a$-DOF manipulator mounted onto a differential drive mobile base, assuming that the low-level controllers of both robots ensure accurate reproduction of a desired velocity profile. We consider the following extended joint space for the robot:
\begin{align}\label{eq: space_coords}
    \textbf{q} &= \begin{bmatrix}
                    \textbf{q}_b\\ 
                    \textbf{q}_a
               \end{bmatrix}
\end{align}
in which $\textbf{q}_a \in \mathbb{R}^{n_a}$ are the generalized coordinates of the manipulator, while $\textbf{q}_b \in \mathbb{R}^{n_b}$ are the angular position of the wheels of the mobile base. We define as $N = n_a + n_b$ the dimension of the augmented configuration space.

We construct the augmented mapping for the robot
\begin{equation}\label{eq: kin_mapping}
    {}^W\dot{\textbf{x}}_E = \textbf{J}_m \dot{\textbf{q}} 
\end{equation}
relating the end tip velocity w.r.t the world frame ${}^W\dot{\textbf{x}}_E \in \mathbb{R}^m$ with the robot velocity in the augmented joint space. We refer to $\textbf{J}_m \in \mathbb{R}^{m \times N}$ as the augmented Jacobian.

We can obtain the desired velocity input in the joint space $\textbf{u} \in \mathbb{R}^{N}$ which implements a desired task space velocity $\dot{\textbf{x}}_{des}$ as $\textbf{u} = \dot{\textbf{q}}_{des} = J_m^+ \dot{\textbf{x}}_{des}$ in which $J_m^+ \in \mathbb{R}^{N \times m}$ is the Moore-Penrose pseudo-inverse of the augmented Jacobian.

We finally assume that the human-robot contact takes place via a commonly manipulated object, whose interaction force $\textbf{F}_e \in \mathbb{R}^m$ is sensed by an haptic interface. We model the interaction using the whole body admittance controller in Fig.~\ref{fig: adm_scheme}, i.e. by integrating the interaction force via the admittance dynamics, synthesizing the admittance velocity $\dot{\textbf{x}}_a \in \mathbb{R}^m$. We can then ensure the reproduction of the admittance dynamics by setting ${}^W\dot{\textbf{x}}_E \approx \dot{\textbf{x}}_{des} \approx \dot{\textbf{x}}_a$. 

In order to enhance the performance of the controller, we want to vary the admittance parameters according to the perceived intention of the human operator, assisting him during the manipulation of the object at the best of the robot capabilities (see e.g. \cite{ferraguti2019variable}). Passivity is however lost if we consider a variable admittance controller, i.e. whose dynamic parameters vary over time. Consider the following time-varying admittance model
\begin{equation}\label{eq: adm_tvar_model}
    \textbf{M}(\textbf{x}_a,t)\Ddot{\textbf{x}}_a + \textbf{C}(\textbf{x}_a, \dot{\textbf{x}}_a, t)\dot{\textbf{x}}_a + \textbf{D}(\textbf{x}_a, t)\dot{\textbf{x}}_a + \frac{\partial P}{\partial \textbf{x}_a}(t) = \textbf{F}_e
\end{equation}
where $\textbf{M}(\textbf{x}_a) = \textbf{M}^T(\textbf{x}_a) \geq 0$ is the inertia matrix, $\textbf{C}(\textbf{x}_a, \dot{\textbf{x}}_a)$ represents the Coriolis term, $\textbf{D}(\textbf{x}_a) \geq 0$ is a damping matrix and $P : \mathbb{R}^m \rightarrow \mathbb{R}$ is a potential field acting on the system.
Here, the variation of the dynamic parameters might end up introducing energy into the system, thus threatening passivity (see e.g. \cite{secchi2019energy, ferraguti2015energy}). The standard admittance controller cannot in fact ensure a robustly stable interaction when reproducing a time-varying dynamics.

Alongside passivity, other conditions need to be satisfied for ensuring safety during the transportation task, both for the human, the robot and the manipulated object. These conditions require a dedicated proper formulation.

We aim at realising a control architecture capable of performing a collaborative human-robot transportation task in a safe and flexible way. The architecture must be capable of passively implementing a variable admittance dynamics as in \eqref{eq: adm_tvar_model}, whose parameters must adapt to properly assist the human operator, alongside accomplishing multiple safety-related and application-oriented tasks, while treating the mobile manipulator as a single whole-body system.

\section{Whole Body Kinematic Model} \label{sec: kin_model}
%\begin{itemize}
    %\item Construction of the Whole-Body kinematic model of the mobile manipulator
    %\item Jacobian of mobile base
    %\item IO-SFL
    %\item Jacobian of the arm
    %\item Mapping of the velocities of the arm
    %\item Mapping of the velocities of the base onto the arm
    %\item Final augmented Jacobian
%\end{itemize}
In this section we present the construction of the augmented kinematic mapping in \eqref{eq: kin_mapping}.

First, we separately compute the kinematic contribution of the arm to the end-effector velocity ${}^{W}\dot{\textbf{x}}_{E(a)}$ and the contribution of the mobile base ${}^{W}\dot{\textbf{x}}_{E(b)}$.
Considering the kinematic mapping of the arm, this is readily available as:
\begin{equation}\label{eq: kin_mapp}
    {}^{0}\dot{\textbf{x}}_{a} = \textbf{J}_{a} \dot{\textbf{q}}_a
\end{equation}
being $\textbf{J}_{a}$ the geometric Jacobian of the arm, relating the velocity of the terminating link $\mathcal{F}_{a}$ w.r.t the initial link frame $\mathcal{F}_0$ expressed in $\mathcal{F}_{0}$ with the joint velocities $\dot{\textbf{q}}_a$. We assume that the tool frame is related to the end tip one via a constant homogeneous transformation matrix ${}^{a}\textbf{T}_E$, as the initial frame is related to the world frame via ${}^{W}\textbf{T}_0$. The contribution of the manipulator to the end-effector velocity, expressed in the world frame, is obtained as:
\begin{align}\label{eq: arm_to_EE}
    \begin{split}
    {}^{W}\dot{\textbf{x}}_{E(a)} = \textbf{J}_A \dot{\textbf{q}}_a \\
    \textbf{J}_A = {}^W\mathcal{T}_0 {}^E\mathcal{T}_{a} \textbf{J}_{a}
    \end{split}
\end{align}
in which ${}^W\mathcal{T}_0$ and ${}^E\mathcal{T}_{a}$ are the adjoint matrices:
\begin{align}\label{eq: spatial_trans_mat1}
    {}^E\mathcal{T}_{a} &= \begin{bmatrix}
                                {}^{a}\textbf{R}_E^T& \quad -{}^{a}\textbf{R}_E^T \times {}^{a}\textbf{r}_E\\ 
                                \quad \textbf{0}_3 &\quad {}^{a}\textbf{R}_E^T
               \end{bmatrix}
\end{align}

\begin{align} \label{eq: spatial_trans_mat2}
    {}^W\mathcal{T}_{0} &= \begin{bmatrix}
                                {}^{W}\textbf{R}_0& \quad
                                {}^{W}\textbf{r}_0 \times
                                {}^{W}\textbf{R}_0\\
                                \quad \textbf{0}_3 &\quad {}^{W}\textbf{R}_0
               \end{bmatrix}
\end{align}
in which ${}^{a}\textbf{R}_E \in \mathbb{SO}(3)$ is the rotation matrix obtained from ${}^a\textbf{T}_{E}$, with ${}^{a}\textbf{r}_E \in \mathbb{R}^3$ being the position vector of the transformation. Similarly, we can extract ${}^{W}\textbf{R}_0 \in \mathbb{SO}(3)$ and ${}^{W}\textbf{r}_0$ from ${}^W\textbf{T}_0$.% = {}^W\textbf{T}_0 {}^0\textbf{T}_{a} {}^{a}\textbf{T}_E$.

For the mobile base, we utilise the kinematic mapping commonly found in literature for a differential drive robot (see e.g. \cite{lavalle_2006}). Given the choice of $\textbf{q}_b$, we indicate as $\dot{\textbf{q}}_b = [\omega_l \,\, \omega_r ]^T$ the angular velocity of the left and right wheel respectively. We can then define the kinematic mapping of the mobile base as:
\begin{equation}\label{eq: kin_map_base}
    {}^{W}\dot{\textbf{x}}_{b} = \textbf{J}_b \dot{\textbf{q}}_b
\end{equation}
which relates the velocity of the wheels with the Cartesian velocity of the mobile base w.r.t the world frame $\mathcal{F}_W$ expressed in $\mathcal{F}_W$. The Jacobian $\textbf{J}_b \in \mathbb{R}^{m \times n_b}$ stems from the kinematic model of the unicycle together with the geometric parameters of the differential drive system:
\begin{align} \label{eq: jac_diff_drive}
    \textbf{J}_b = \begin{bmatrix*}[c]
                   \frac{r\,\cos(\theta)}{2} & \frac{r\,\cos(\theta)}{2} \\
                   \frac{r\,\sin(\theta)}{2} & \frac{r\,\sin(\theta)}{2} \\
                   \multicolumn{2}{c}{\textbf{0}_{3 \times 2}} \\
                   \frac{r}{L} & -\frac{r}{L}
                   \end{bmatrix*}
\end{align}
in which $L$ is the distance between the two wheels, $r$ is the radius of the wheels and $\theta$ is the steering angle of the mobile base w.r.t. the world frame $\mathcal{F}_W$.

The kinematic model in \eqref{eq: kin_map_base} considers as a reference point for the motion the midpoint of the wheel axis. This is however subject to non-holonomic constraints, which limit the instantaneous mobility of the robot. In order to overcome this limitation, we deploy a I-O SFL controller \cite{d1995control}, which allows us to linearize static state feedback laws. By shifting to a reference point located at a distance $b$ from the wheel axis w.r.t. the base frame $\mathcal{F}_b$, \fb{the translational
velocity of the reference control point is unrestricted}. The kinematic model in \eqref{eq: kin_map_base} becomes:
\begin{equation} \label{eq: kin_map_IOSFL}
    {}^{W}\dot{\textbf{x}}_{B} = \textbf{J}_B \dot{\textbf{q}}_b
\end{equation}
in which ${}^{W}\dot{\textbf{x}}_{B}$ is the Cartesian velocity of the new control point w.r.t the world frame $\mathcal{F}_W$, while the corresponding Jacobian $\textbf{J}_B$ is defined as:
\begin{align} \label{eq: jac_iosfl}
    \textbf{J}_B = \begin{bmatrix*}[c]
                   (\frac{r\,\cos\theta}{2} - \frac{br\,\sin(\theta)}{L}) & (\frac{r\,\cos\theta}{2} + \frac{br\,\sin(\theta)}{L}) \\
                   (\frac{r\,\sin\theta}{2} + \frac{br\,\cos(\theta)}{L}) & (\frac{r\,\sin\theta}{2} - \frac{br\,\cos(\theta)}{L}) \\
                   \multicolumn{2}{c}{\textbf{0}_{3 \times 2}} \\
                   \frac{r}{L} & -\frac{r}{L}
                   \end{bmatrix*}
\end{align}
Finally, we introduce a mapping term which allows us to compute the contribution to the end effector velocity due to the mobile base as:
\begin{equation}\label{eq: kin_map_base2EE}
    {}^{W}\dot{\textbf{x}}_{E(b)} = \mathbb{H}\,\textbf{J}_B \dot{\textbf{q}}_b
\end{equation}
in which $\mathbb{H} \in \mathbb{R}^{m \times m}$ is defined as:
\begin{align}
    \mathbb{H} =   \begin{bmatrix*}[c]
                   1 & 0 & 0 & 0 & 0 &-{}^{B}y_E \\
                   0 & 1 & 0 & 0 & 0 & \phantom{-} {}^{B}x_E \\
                   \multicolumn{6}{c}{\textbf{0}_{3 \times 6}} \\
                   0 & 0 & 0 & 0 & 0 & \phantom{-}1
                   \end{bmatrix*}
\end{align}
with $({}^{B}x_E,{}^{B}y_E)$ being the Cartesian coordinates of the end effector with respect to the frame $\mathcal{F}_B$.

Merging \eqref{eq: arm_to_EE} and \eqref{eq: kin_map_base2EE} we can achieve the mapping in \eqref{eq: kin_mapping} using the augmented generalized velocities $\dot{\textbf{q}} = [\dot{\textbf{q}}_b^T \,\,\dot{\textbf{q}}_a^T]^T$
and by constructing the augmented Jacobian $\textbf{J}_m$ as:
\begin{align} \label{eq: augm_jacobian}
    \textbf{J}_m = \begin{bmatrix}
                            \mathbb{H}\,\textbf{J}_B \quad \textbf{J}_A
                       \end{bmatrix}
\end{align}
We can thus track a desired reference velocity for the robot $\dot{\textbf{x}}_{des} = {}^{W}\dot{\textbf{x}}_{E(a)} + {}^{W}\dot{\textbf{x}}_{E(b)}$ via the generalized velocities:
\begin{equation} \label{eq: inverse_augm_kin}
    \dot{\textbf{q}}_{des} = \textbf{J}_m^+ \dot{\textbf{x}}_{des} 
\end{equation}

\section{Collaborative Constraint-Oriented Control Architecture} \label{sec: tanks}
In this section we provide the necessary tools for building the constraint-oriented control architecture proposed in \cite{benzi2021optimization}, which is based on energy tanks \cite{franken2011bilateral} and CBFs for task encoding \cite{notomista2019constraint}. We finally present the policy for adapting the admittance parameters online. %We first present energy tanks \cite{ferraguti2015energy},\cite{franken2011bilateral} and the tank-based formulation of a variable admittance controller \eqref{eq: adm_tvar_model}. Then, we introduce Control Barrier Functions and their exploitation for encoding time-varying tasks \cite{notomista2019constraint,notomista2019optimal}. We then merge everything into a single optimization problem. Finally, we present the policy for adapting the admittance parameters online.
\subsection{Energy Tanks}
Energy tanks are energy reservoirs modelled as:
\begin{equation} \label{eq: tank_model}
    \begin{cases}
        \dot{x}_{t} = u_{t}\\
        y_{t} = \frac{\partial{T}}{\partial{x_{t}}} = x_{t}(t)
    \end{cases}
\end{equation}
where $x_{t}\in \mathbb{R}$ is the state of the tank, the pair $(u_{t},y_{t}) \in\mathbb{R}\times\mathbb{R}$ represents the power port through which the tank exchanges energy and
\begin{equation} \label{eq: tank_energy}
    T(x_t)= \frac{1}{2}x^{2}_{t}
\end{equation}
is the storage function of the energy tank.

The energy stored inside tanks is not affected by any specific dynamics, meaning that it can be purposefully utilised at any stage for performing any desired behaviour.
This is carried out, in the case of the standard admittance controller in \eqref{eq: adm_tvar_model}, by interconnecting the power port of the tank $(u_t,y_t)$ with the power port $(\textbf{F}_e,\dot{\textbf{x}}_{des})$ of the implemented admittance dynamics, via:
% \begin{equation} \label{eq: tank_modulation}
%     \begin{cases}
%     u_{t}(t) = \textbf{A}^{T}(t)\textbf{F}_e(t)\\
%     \dot{\textbf{x}}_{des}(t) = \textbf{A}(t)y_{t}(t)
%     \end{cases}
% \end{equation}
\begin{equation} \label{eq:tank modulated description}
    \begin{cases}
        u_t(t) = \dot{x}_{t}(t) = \textbf{A}^{T}(t)\textbf{F}_e(t)\\
        \dot{\textbf{x}}_{des}(t) = \textbf{A}(t)y_{t}(t)= \bm{\gamma}(t)
    \end{cases}
\end{equation}

where $\textbf{A}(t) \in \mathbb{R}^{m}$ is defined as
\begin{equation} \label{eq:modulation matrix description}
    \textbf{A}(t) = \frac{\bm{\gamma}(t)}{x_{t}(t)}
\end{equation}
and $\bm{\gamma}(t) \in \mathbb{R}^{n}$ is the desired value for the output $\dot{\textbf{x}}_{des}(t)$, i.e. the velocity resulting from the admittance dynamics. This means that by a proper modulation of the energy flow it is possible to implement any desired port behaviour. From \eqref{eq: tank_model},\eqref{eq: tank_energy} and \eqref{eq:tank modulated description} we obtain
\begin{equation}\label{eq:tankadmbalance}
\dot{T}=u_t y_t=\textbf{A}^T (t)\textbf{F}_e y_t=\bm{\gamma}^T \textbf{F}_e
\end{equation}
showing how the energy flow necessary for implementing a desired admittance dynamics is provided by the tank.

A singularity occurs in \eqref{eq:tank modulated description} whenever $x_{t}(t)=0$ due to the definition of $\textbf{A}(t)$, meaning that the tank is depleted and no behavior can be performed. This issue can be faced by initializing $x_{t}$ such that $T(x_{t}(0)) \geq \underline \varepsilon > 0$ and by guaranteeing that $T(x_{t}(t)) \geq \underline \varepsilon$ $\forall t>0$, with $\underline \varepsilon$ being an arbitrarily set lower bound.

In \cite{secchi2019energy}, it was shown that if $T(x_{t})\geq \underline{\varepsilon}\,\, \forall t\geq0$, then the modulated tank \eqref{eq:tank modulated description} remains passive independently of the desired output $\bm{\gamma}(t)$.
% \newtheorem{prop}{Proposition}
% \begin{prop} \label{prop: always passive}
 %	If $T(x_{t})\geq \underline{\varepsilon}$ for all $t\geq0$, then the modulated tank \eqref{eq:tank modulated description} remains passive independently of the desired output $\bm{\gamma}(t)$.
 %\end{prop}
Thus, as long as the tank is not depleted, any desired port behavior can be passively implemented by modulating the energy flow. Specifically, it is possible to reproduce any passive dynamics \cite{giordano2013IJRR, riggio2018use}.

For reproducing a non-passive dynamics, such as the time varying admittance in \eqref{eq: adm_tvar_model}, we can exploit the previous results for passivizing the desired behaviour. The passivity of the modulated tank can be encoded as a constraint:
 \begin{equation}\label{eq: passivity constraint}
     T(x_{t}) \geq \underline{\varepsilon} \quad \forall t\geq 0
 \end{equation}
Then, as in \cite{secchi2019energy} and \cite{benzi2021optimization}, the appropriate control input can be synthesised via the following optimization problem
 \begin{equation} \label{eq: opt problem ICRA19}
    \begin{aligned}
        & \underset{\dot{\textbf{x}}_{des}}{\text{minimize}}
        & & ||\dot{\textbf{x}}_{des} - \dot{\textbf{x}}_{a}||^{2} \\
        & \text{subject to}
        & & \int_{0}^{t}\textbf{F}_{e}^{T}(\tau)\dot{\textbf{x}}_{des}(\tau)d\tau\geq - T(x_{t}(0)) + \underline{\varepsilon}
    \end{aligned}
\end{equation}
 The solution of \eqref{eq: opt problem ICRA19} provides the best passive approximation of the desired behavior $\dot{x}_{a}$, by keeping track of the energy stored in the tank. We then set $\dot{\textbf{x}}_{des} = \bm{\gamma}$ and utilize it to tune the modulation matrix $A(t)$ in \eqref{eq:tank modulated description}. This ensures a passive energy balance even if the variation of the parameters would inject energy in the system.

 A discrete-time version of \eqref{eq: opt problem ICRA19} was proposed in \cite{secchi2019energy}, thus obtaining a convex formulation, for a  computationally fast and simple optimization problem. %These features allow for real-time execution and for a seamless addition into a larger optimization-based framework.
 \subsection{Control Barrier Functions}
 CBFs can be formulated for enforcing multiple time varying constraints on a robot described by \eqref{eq: kin_mapping}. A specific procedure was described in \cite{notomista2020set}, by which both a set of tasks, both kinematic limits are modelled as dynamic constraints onto the input of the systems. These can then be inserted into an optimization problem, whose solution is the input satisfying all the constraints, i.e. leading to the simultaneous execution of all the tasks.

We represent the desired tasks as the minimization of a non-negative, time-varying, continuously-differentiable cost function $C : \mathbb{R}^{n} \times \mathbb{R} \rightarrow \mathbb{R}$. Considering a robot modelled as \eqref{eq: kin_mapping} and the time-varying task variable $\bm{\sigma} \in \mathbb{R}^{n}$ as an output variable, the task execution can be encoded via the following optimization problem:
\begin{equation} \label{eq: prob cost function}
    \begin{aligned}
    & \underset{u}{\text{minimize}}
    & & C(\bm{\sigma},t) \\
    & \text{subject to}
    & & {}^{W}\dot{\textbf{x}}_E = \textbf{J}_m (\textbf{q})\,\textbf{u} \\
    &&& \bm{\sigma} = \textbf{k}({}^{W}\textbf{x}_E,t)
    \end{aligned}
\end{equation}
\fb{A convex formulation of the problem can be obtained by leveraging Control Barrier Functions.
Let $\mathcal{C}\subset \mathbb{R}^n$ be the subset in which the task is considered to be executed, i.e. $C(\bm{\sigma},t)=0$. Let then $h:\mathbb{R}^{n} \times \mathbb{R} \rightarrow \mathbb{R}$ be a control barrier function defined as $h(\bm{\sigma}, t)=-C(\bm{\sigma},t)$. We have that $h$ is non negative only whenever $C(\bm{\sigma},t)=0$, i.e. the region of satisfaction of the task. By enforcing the non-negativity of $h$ we can then achieve the execution of the task $\bm{\sigma}$}. This results in the following convex optimization problem \cite{notomista2019constraint}:
\begin{equation} \label{eq: prob CBF general}
    \begin{aligned}
    & \underset{\dot{\textbf{q}}}{\text{minimize}}
    & & ||\dot{\textbf{q}}||^{2} \\
    & \text{subject to}
    & & \frac{\partial h}{\partial t} + \frac{\partial h}{\partial \bm{\sigma}}\frac{\partial \bm{\sigma}}{\partial {}^{W}\textbf{x}_E}\textbf{J}_m(\textbf{q})\dot{\textbf{q}} + \alpha(h(\bm{\sigma},t)) \geq 0
    \end{aligned}
\end{equation}
where $\alpha(\cdot)$ is an extended class $\mathcal{K}$ function\footnote{An extended class $\mathcal{K}$ is a function $\phi:\mathbb{R}\rightarrow\mathbb{R}$ such that $\phi$ is strictly increasing and $\phi (0)=0$} and where we have chosen as an input for \eqref{eq: kin_mapping} $\textbf{u}=\dot{\textbf{q}}$. 

We can straightfowardly extended the formulation in order to implement the execution of multiple M different tasks at the same time, each respectively encoded by the cost functions $C_{1}, \dots, C_{M}$. %The symultaneous execution of these tasks can then be achieved by solving the following convex optimization problem:
% \begin{equation} \label{eq: prob CBF multitask slack}
%     \begin{aligned}
%     & \underset{\dot{\textbf{q}}, \bm{\delta}}{\text{minimize}}
%     & & ||\dot{\textbf{q}}||^{2} + l||\bm{\delta}||^{2} \\
%     & \text{subject to}
%     & & \frac{\partial h_{m}}{\partial t} + \frac{\partial h_{m}}{\partial \bm{\sigma}}\frac{\partial \bm{\sigma}}{\partial {}^{b}\textbf{x}_E}\textbf{J}_m(\textbf{q})\dot{\textbf{q}} \\
%     &&& + \alpha (h_{m}(\bm{\sigma},t)) \geq -\bm{\delta}_{m} \quad  m \in  \{ 1, \dots, M\}
%     \end{aligned}
% \end{equation}
% in which $h_{m}(\bm{\sigma},t) = -C_{m}(\bm{\sigma},t)$ and $\bm{\delta} = [\bm{\delta}_{1},\dots,\bm{\delta}_{M}]^{T}$ is the vector of slack variables dedicated to relaxing each constraint, while $l \geq 0$ is a scaling factor. The slack variables $\bm{\delta}_{i}$ are required for ensuring the feasibility of the problem at all times, even if conflicting constraints are simultaneously active.
At the same time, we can add the passivity constraint in \eqref{eq: opt problem ICRA19} to the stack, thus obtaining a single convex optimization problem as in \cite{benzi2021optimization}. The final formulation is:
\begin{equation} \label{eq: prob CBF multitask slack prio passive adm}
    \begin{aligned}
    & \underset{\dot{\textbf{q}}, \bm{\delta}}{\text{minimize}}
    & & ||\dot{\textbf{q}} - \dot{\textbf{q}}_{adm}||^{2} + l||\bm{\delta}||^{2} \\
    & \text{subject to}
    & & \frac{\partial h_{m}}{\partial t} + \frac{\partial h_{m}}{\partial \bm{\sigma}}\frac{\partial \bm{\sigma}}{\partial {}^{W}\textbf{x}_E}\textbf{J}_m(\textbf{q})\dot{\textbf{q}} \\
    &&& + \alpha(h_{m}(\bm{\sigma},t)) \geq -\bm{\delta}_{m} \quad  m \in  \{ 1, \dots, M\}\\
    &&& \int_{0}^{t}\textbf{F}_{e}^{T}(\tau)\textbf{J}_m(\textbf{q})\dot{\textbf{q}}(\tau)d\tau \geq -T(x_{t}(0)) + \underline{\varepsilon}\\
    \end{aligned}
\end{equation}
in which $\dot{\textbf{q}}_{adm} = \textbf{J}_m(\textbf{q})^{+}\dot{\textbf{x}}_{a}$ is the desired admittance expressed in the joint space,  $h_{m}(\bm{\sigma},t) = -C_{m}(\bm{\sigma},t)$ and $\bm{\delta} = [\delta_{1},\dots,\delta_{M}]^{T}$ is the vector of slack variables dedicated to relaxing each constraint, while $l \geq 0$ is a scaling factor. The slack variables $\delta_{m}$ are required for ensuring the feasibility of the problem at all times, even if conflicting constraints are simultaneously active.

 We have thus built an architecture capable of implementing both a passive time-varying admittance controller, both the execution of multiple tasks at the same time.
\subsection{Admittance Parameters Adaptation}
The choice of the dynamic parameters for a given admittance directly affects how the human perceives the interaction and the overall quality of the collaboration \cite{duchaine2007general}.

Our goal is to adapt online the admittance parameters according to his perceived intention of the operator, in order to properly assist him during the collaborative transportation task. We utilize the adaptation policy proposed in \cite{lecours2012variable}, where the human intention is inferred by monitoring the magnitude and the direction of the desired acceleration, alongside the current velocity value. Two possible intentions are taken into account: acceleration and deceleration.
%\begin{itemize}
    %\item \textit{Acceleration:} in which the operator is currently moving in one direction and intends to keep on moving in that direction. This is detected whenever the desired acceleration and the current velocity share the same direction. In this case, to ease the transportation for the human, both the damping and the inertia are lowered according to the magnitude of the acceleration.
    %\item \textit{Deceleration:} in which the operator wants to decelerate or stop the robot motion. Such intention is perceived whenever the acceleration and the velocity present opposite signs. In this case, to reduce the effort for the operator, the damping is increased while the inertia is lowered.
%\end{itemize}
The damping and inertia parameters are updated at each cycle according to the perceived intention, starting from default values of $\textbf{D}_f$ and $\textbf{M}_f$. %Initially, default values for inertia $\textbf{M}_f$ and damping $\textbf{D}_f$ are set. These values will be employed whenever the desired acceleration is sufficiently low or null, i.e. when the operator wants to execute smooth and precise movements. 
\fb{
The damping term $\textbf{D}$ is tuned as follows:
\begin{equation}\label{eq: damping_adapt}
    \begin{cases}
        \textbf{D} = \textbf{D}_f - \mathbb{I}_m \odot (\alpha_a |\Ddot{\textbf{x}}_a| \textbf{1}_m^T) \quad \text{if acceleration} \\
        \textbf{D} = \textbf{D}_f + \mathbb{I}_m \odot (\alpha_d |\Ddot{\textbf{x}}_a| \textbf{1}_m^T) \quad \text{if deceleration}
    \end{cases}
\end{equation}
in which $\alpha_a$ and $\alpha_d$ are two gains to be tuned, $|\cdot|$ indicates the magnitude of the vector, $\mathbb{I}_m \in \mathbb{R}^{m \times m}$ is the identity matrix, $\textbf{1}_m \in \mathbb{R}^m$ is the vector of ones, while $\odot$ indicates the Hadamard (or element-wise) product.
}
The inertia values are computed according to the damping ones, by ensuring that the ratio $\frac{\textbf{M}(\textbf{x}_a)}{\textbf{D}(\textbf{x}_a)}$ remains proportional, \fb{ resulting in a more intuitive interaction for the operator:
\begin{equation}\label{eq: inertia_adapt}
    \begin{cases}
        \textbf{M} = \frac{\textbf{M}\,\textbf{D}}{\textbf{D}_f} \quad &\text{if acceleration} \\
        \textbf{M} = \frac{\textbf{M}}{\textbf{D}_f}(1 - \beta(1 - e^{-\eta(\textbf{D} - \textbf{D}_f)}))\textbf{D} \quad &\text{if deceleration}
    \end{cases}
\end{equation}
in which $\beta \in (0,1)$ serves to tune the steady-state value of $\frac{\textbf{M}(\textbf{x}_a)}{\textbf{D}(\textbf{x}_a)}$, while $\eta$ is a parameter defining the smoothness with which the ratio varies (see \cite{lecours2012variable}).
}
\section{Experiments} \label{sec: experiments}
\begin{figure}[tb]
    \centering
    \includegraphics[height=6cm]{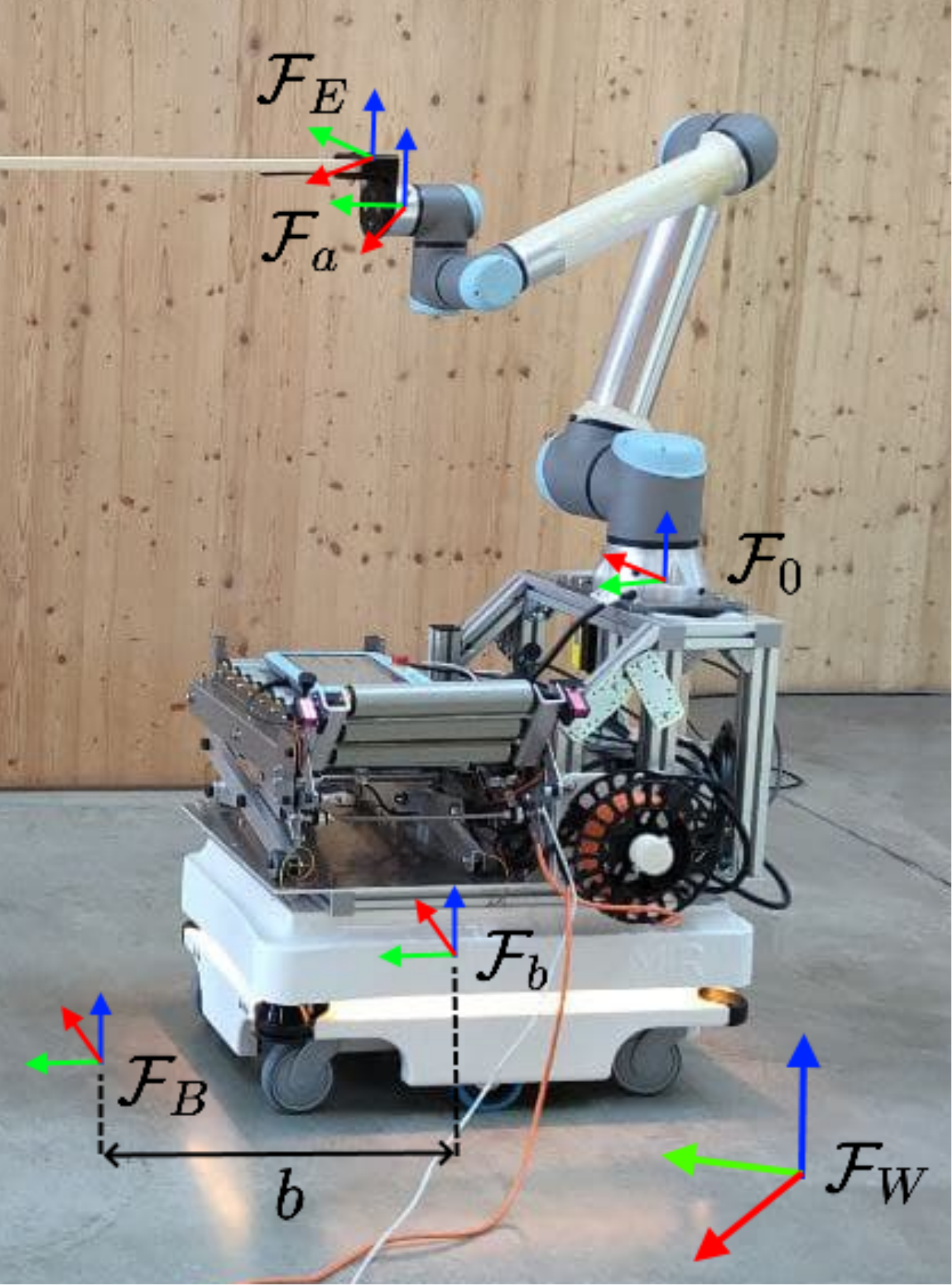}
    \caption{Setup and structure of the mobile manipulator and frames utilized for the kinematic computations}
    \label{fig: frames_and_mm}
\end{figure}
\fb{An experimental evaluation has been conducted, in order to validate each individual component of the architecture. The mobile manipulator used in the validation is composed of a MiR 100 mobile base, with a 6-DOF collaborative manipulator UR10e mounted on top.} In Fig.~\ref{fig: frames_and_mm} the overall setup, as well as the utilized frames can be observed.

\subsection{Passive Time-Varying Admittance Controller}
\begin{figure}[tb]
    \centering
    \includegraphics[height=4cm, width=0.85\columnwidth]{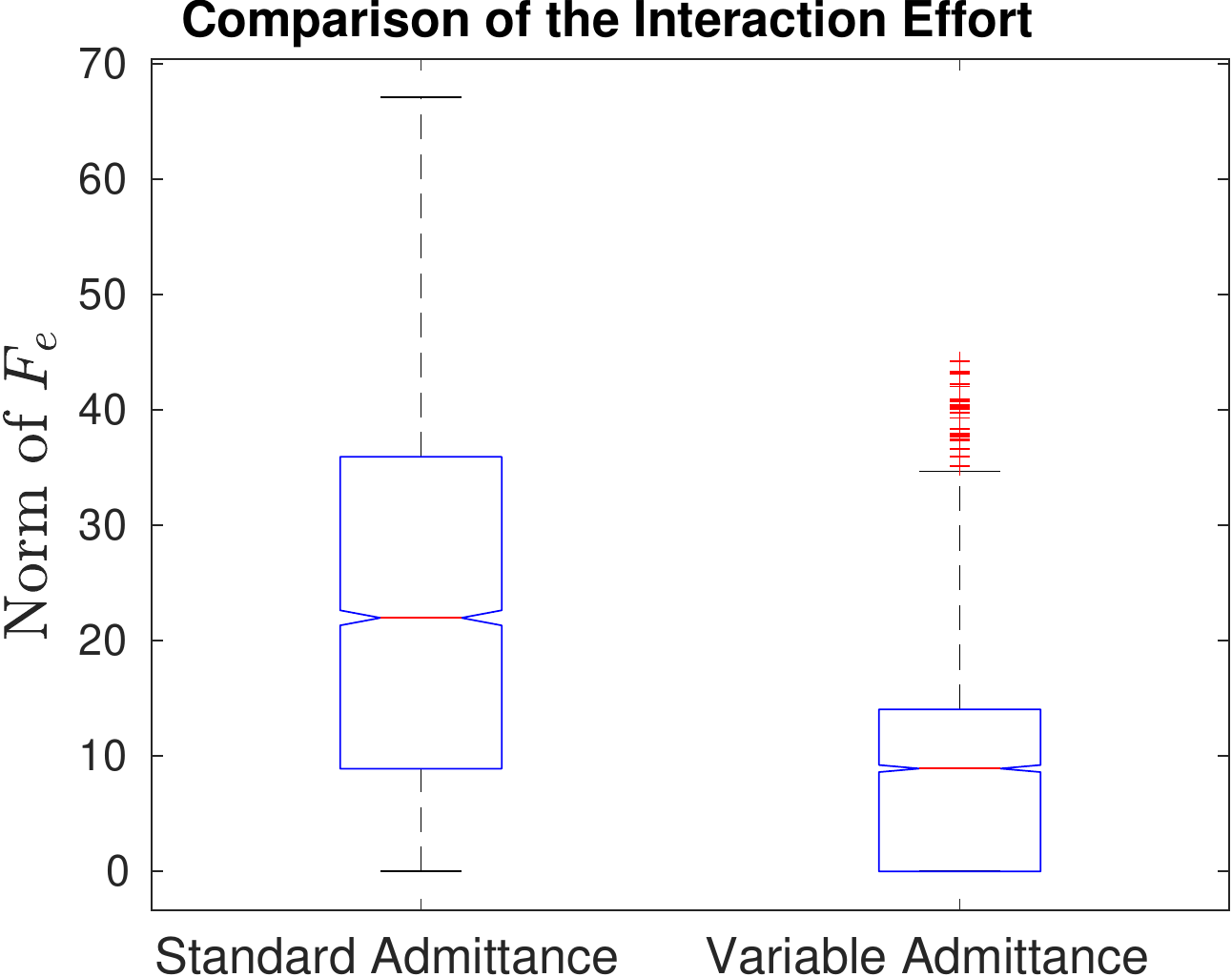}
    \caption{\fb{Distribution over time of the force exerted by the human using the two admittance controllers}}
    \label{fig: effort_confront}
\end{figure}
\begin{figure}[tb]
    \centering
    \includegraphics[height=4cm, width=0.85\columnwidth]{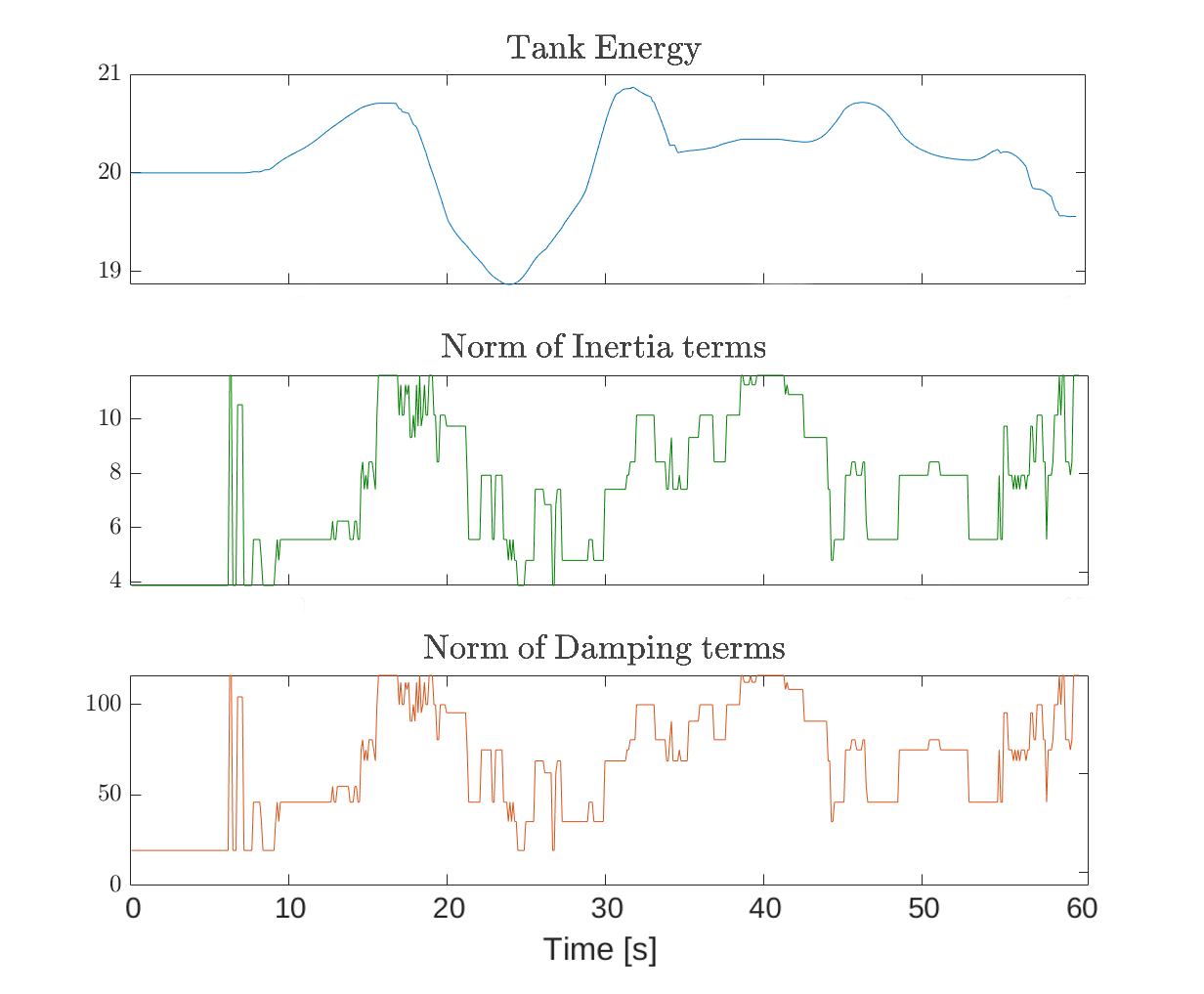}
    \caption{Norm of the variation of the admittance parameters and evolution of the energy in the tank over time}
    \label{fig: dyn_params_vs_tank}
\end{figure}
First, the collaborative transportation task is carried out, highlighting the advantages of the passive time-varying admittance controller. During the experiment, a wooden plank of $2.5m$ in length is attached to the tool of the manipulator (see Fig.~\ref{fig: frames_and_mm}), while the human is holding the opposite side. As the human pulls his side of the plank, the interaction force $\textbf{F}_e$ is sensed by the on-board F/T sensor of the UR10e. After being transformed by means of ${}^W\mathcal{T}_E$ into the world frame, the force is integrated via the time-varying admittance dynamics \eqref{eq: adm_tvar_model} for producing a desired operational velocity set-point $\dot{\textbf{x}}_{des}$. By means of the whole body kinematic mapping in \eqref{eq: inverse_augm_kin}, a desired velocity for both the arm joints and the mobile base wheels is then synthesized. In this way, the robot can actively assist the human in the transportation of the load.

\fb{During the transportation task, an experienced user has to navigate in a constricting environment, with frequent stops and turns. In this scenario, the online adaptation of the admittance parameters allows to visibly enhance the quality of the interaction. In fact, by dynamically varying the mass and damping terms as shown in Sec.~\ref{sec: tanks}, the human is assisted while decelerating and accelerating the robot, resulting in reduced physical effort from his side. This can be observed in Fig.~\ref{fig: effort_confront}, in which the distribution of the force exerted by the operator over time using the standard and the variable admittance controller is portrayed. For the standard case, the values of inertia and damping were fixed at the default ones $\textbf{M}_f$ and $\textbf{D}_f$, namely 4$kg$ and 20$\frac{Ns}{m}$.}

\fb{As shown in Sec.~\ref{sec: tanks}, the overall passivity is ensured by the energy tank and the energetic condition in \eqref{eq: opt problem ICRA19}, thus ensuring a robustly stable interaction. The variation of the dynamic parameters during the task, as well as the evolution of the energy in the tank, are portrayed in Fig.~\ref{fig: dyn_params_vs_tank}.}

\subsection{CBF-based tasks}

\begin{figure}[tb]
    \centering
    \includegraphics[height=4cm, width=0.85\columnwidth]{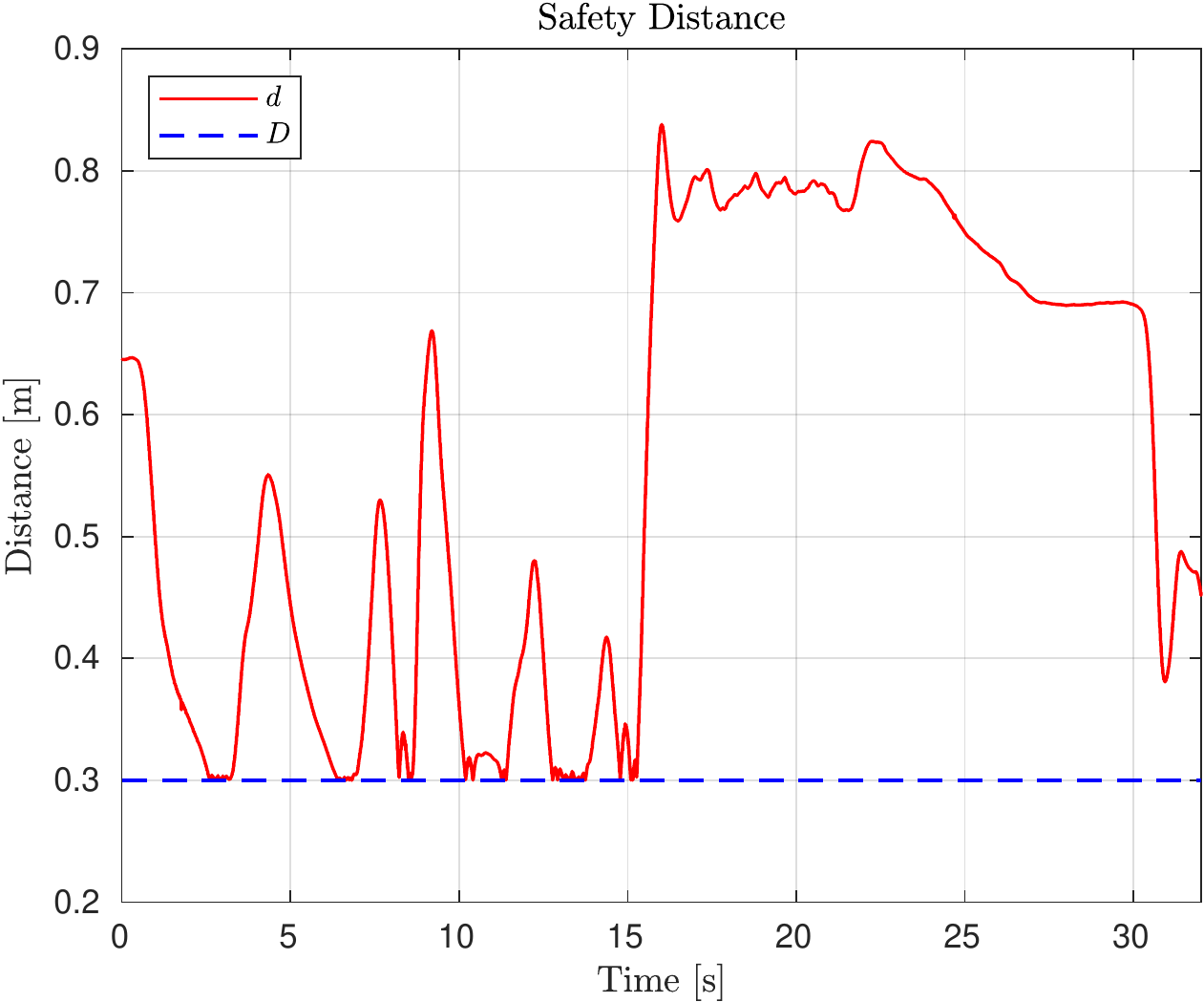}
    \caption{Evolution of the safety distance over time.}
    \label{fig: h_safe}
\end{figure}

\begin{figure}[tb]
    \centering
    \includegraphics[height=4cm, width=0.85\columnwidth]{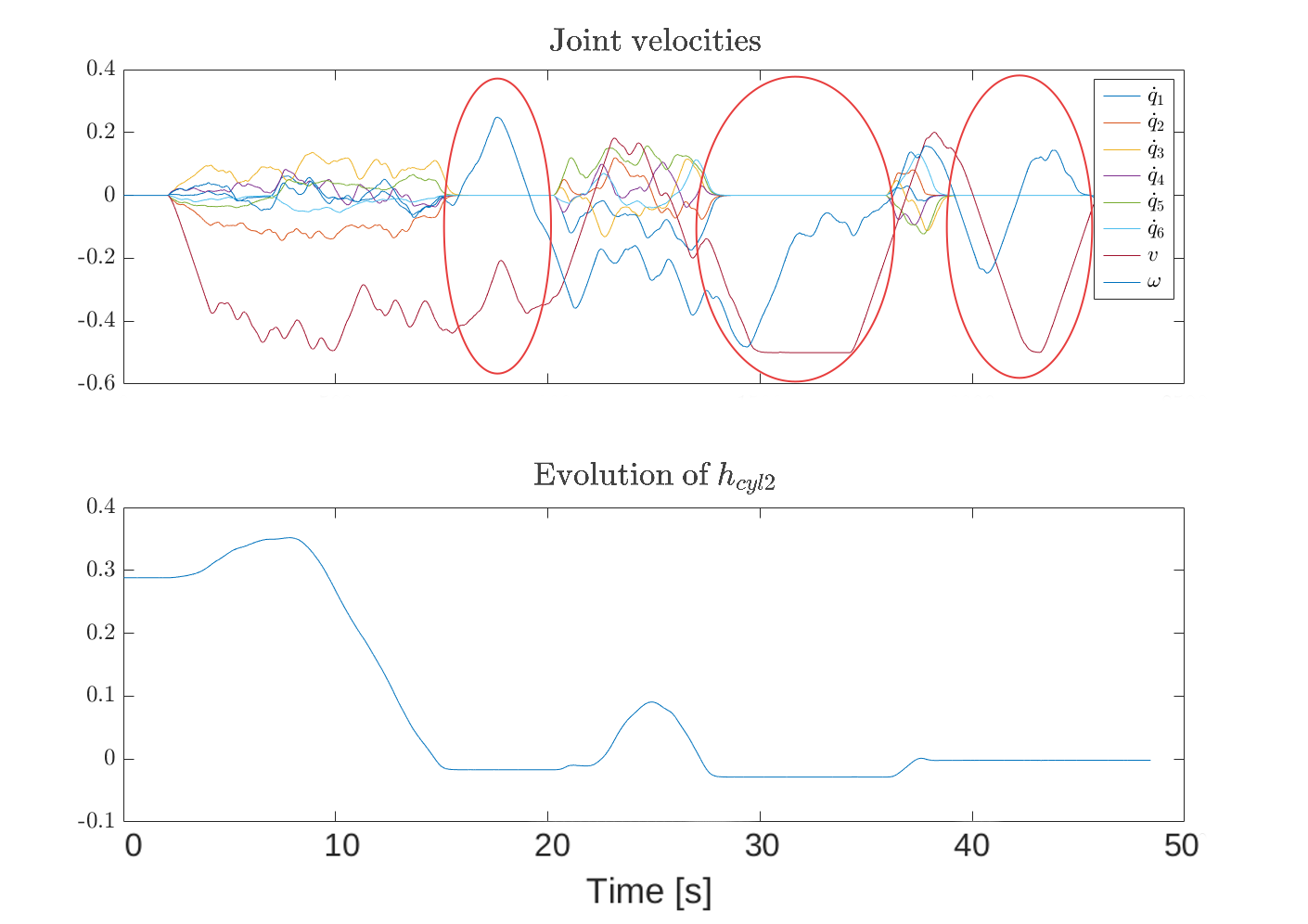}
    \caption{Value of the augmented joint velocities and evolution of $h_{cyl2}$ over time. Whenever $h_{cyl2} = 0$, only the mobile base contributes to the motion.}
    \label{fig: h_cyl2_vs_vel}
\end{figure}

Alongside the main transportation task, a set of additional secondary task is implemented. These are encoded via CBFs and inserted into the optimization problem in \eqref{eq: prob CBF multitask slack prio passive adm}.

First, the safety of the transported object is ensured. The load is wrapped with a capsule \cite{lin2017real}, a virtual object composed by
two semi-spheres, centred in the two mid-points of the short edges of the plank, and a cylinder whose longitudinal axis links the two points. Then, a minimum safety distance $D_{min} = 0.3\,m$ between the capsule and a detectable object in the scene is guaranteed by means of the following CBF
\begin{equation}\label{eq: safety_CBF}
    h_{safe} = d^2 - D^2_{min}
\end{equation}
in which the safety distance $d \in \mathbb{R}$ is defined as
\begin{equation}\label{eq: capsule_distance}
    d = || \textbf{x}_{obs} - \textbf{x}_c || - r_c
\end{equation}
where $\textbf{x}_{obs} \in \mathbb{R}^m$ is the position of the obstacle, while $\textbf{x}_c \in \mathbb{R}^m$ is the position on the capsule which is closest to the obstacle and $r_c \in \mathbb{R}$ is the radius of the capsule. Given that the manipulator is significantly more dexterous than the mobile base and that sudden rotations of the base can be negatively perceived by the human, we decided to only employ the manipulator for this safety task. Thus, its execution can performed by adding in \eqref{eq: prob CBF multitask slack prio passive adm} the constraint:
\begin{equation}\label{eq: safety_capsule_constraint}
    2(\textbf{x}_{obs} - \textbf{x}_c)^T \textbf{J}_{lim} \dot{\textbf{q}} \geq  -\alpha (h_{safe}) + \delta_{safe}
\end{equation}
with $\textbf{J}_{lim}= [\textbf{0}_{6 \times 2}^T \,\, \textbf{J}_a^T]^T$ being the restriction of the augmented Jacobian to the manipulator alone.
The plot in Fig~\ref{fig: h_safe} portrays the evolution of $d$ over time, showing how the safety distance is always respected.

This procedure can be extended for guaranteeing the safety of the overall structure by wrapping with capsules the entire system.

Additionally, we restrict the workspace of the manipulator to a suitable subset for the application. First, in order to avoid self collision between the tool mounted onto the manipulator and its joints, we define the following CBF
\begin{equation}\label{eq: h_cyl_1_def}
    h_{cyl1} = d_1^2 - R_1^2
\end{equation}
with $d_1 = \sqrt{{}^0x_{E1}^2 + {}^0x_{E2}^2}$ and $R_1$ being the radius which defines a safety cylinder centred in the manipulator base.

Additionally, we want to limit the outreach of the manipulator, in order to avoid singularities during the transportation, as they could lead to unexpected behaviours. We then deploy the following CBF for forcing the tip of the manipulator to stay inside a set cylinder of radius $R_2$
\begin{equation}\label{eq: h_cyl_2_def}
    h_{cyl2} = -d_1^2 + R_2^2
\end{equation}
Fig.~\ref{fig: h_cyl2_vs_vel} shows the contribution of this CBF; as the manipulator reaches the edge of the cylinder, its joint velocities halt, restarting only as it recoils in the desired area. 
\section{Conclusions and Future Works} \label{sec: conclusions}
In this paper we developed a control architecture capable of performing a flexible and robust collaborative transportation task, treating the mobile manipulator as a whole-body system and considering non-holonomic constraints. In future works, we aim at further exploiting the redundancy of the robot, for maximizing performance and compliance.

\bibliography{ifacconf}

\begin{thebibliography}{28}
\providecommand{\natexlab}[1]{#1}
\providecommand{\url}[1]{\texttt{#1}}
\providecommand{\urlprefix}{URL }
\expandafter\ifx\csname urlstyle\endcsname\relax
  \providecommand{\doi}[1]{doi:\discretionary{}{}{}#1}\else
  \providecommand{\doi}{doi:\discretionary{}{}{}\begingroup
  \urlstyle{rm}\Url}\fi

\bibitem[{Agravante et~al.(2013)Agravante, Cherubini, Bussy, and
  Kheddar}]{agravante2013human}
Agravante, D.J., Cherubini, A., Bussy, A., and Kheddar, A. (2013).
\newblock Human-humanoid joint haptic table carrying task with height
  stabilization using vision.
\newblock In \emph{2013 IEEE/RSJ International Conference on Intelligent Robots
  and Systems}, 4609--4614. IEEE.

\bibitem[{Ames et~al.(2019)Ames, Coogan, Egerstedt, Notomista, Sreenath, and
  Tabuada}]{ames2019control}
Ames, A.D., Coogan, S., Egerstedt, M., Notomista, G., Sreenath, K., and
  Tabuada, P. (2019).
\newblock Control barrier functions: Theory and applications.
\newblock In \emph{2019 18th European Control Conference (ECC)}, 3420--3431.
  IEEE.

\bibitem[{Benzi et~al.(2022)Benzi, Ferraguti, Riggio, and
  Secchi}]{benzi2022shared}
Benzi, F., Ferraguti, F., Riggio, G., and Secchi, C. (2022).
\newblock An energy-based control architecture for shared autonomy.
\newblock \emph{IEEE Transactions on Robotics}, 1--19.
\newblock \doi{10.1109/TRO.2022.3180885}.

\bibitem[{Benzi and Secchi(2021)}]{benzi2021optimization}
Benzi, F. and Secchi, C. (2021).
\newblock An optimization approach for a robust and flexible control in
  collaborative applications.
\newblock In \emph{2021 IEEE International Conference on Robotics and
  Automation (ICRA)}, 3575--3581. IEEE.

\bibitem[{Calinon et~al.(2009)Calinon, Evrard, Gribovskaya, Billard, and
  Kheddar}]{calinon2009learning}
Calinon, S., Evrard, P., Gribovskaya, E., Billard, A., and Kheddar, A. (2009).
\newblock Learning collaborative manipulation tasks by demonstration using a
  haptic interface.
\newblock In \emph{2009 International Conference on Advanced Robotics}, 1--6.
  IEEE.

\bibitem[{Calinon et~al.(2007)Calinon, Guenter, and
  Billard}]{calinon2007learning}
Calinon, S., Guenter, F., and Billard, A. (2007).
\newblock On learning, representing, and generalizing a task in a humanoid
  robot.
\newblock \emph{IEEE Transactions on Systems, Man, and Cybernetics, Part B
  (Cybernetics)}, 37(2), 286--298.

\bibitem[{d'Andr{\'e}a Novel et~al.(1995)d'Andr{\'e}a Novel, Campion, and
  Bastin}]{d1995control}
d'Andr{\'e}a Novel, B., Campion, G., and Bastin, G. (1995).
\newblock Control of nonholonomic wheeled mobile robots by state feedback
  linearization.
\newblock \emph{The International journal of robotics research}, 14(6),
  543--559.

\bibitem[{Dimeas and Aspragathos(2016)}]{dimeas2016online}
Dimeas, F. and Aspragathos, N. (2016).
\newblock Online stability in human-robot cooperation with admittance control.
\newblock \emph{IEEE transactions on haptics}, 9(2), 267--278.

\bibitem[{Duchaine and Gosselin(2007)}]{duchaine2007general}
Duchaine, V. and Gosselin, C.M. (2007).
\newblock General model of human-robot cooperation using a novel velocity based
  variable impedance control.
\newblock In \emph{Second Joint EuroHaptics Conference and Symposium on Haptic
  Interfaces for Virtual Environment and Teleoperator Systems (WHC'07)},
  446--451. IEEE.

\bibitem[{Ferraguti et~al.(2015)Ferraguti, Preda, Manurung, Bonfe, Lambercy,
  Gassert, Muradore, Fiorini, and Secchi}]{ferraguti2015energy}
Ferraguti, F., Preda, N., Manurung, A., Bonfe, M., Lambercy, O., Gassert, R.,
  Muradore, R., Fiorini, P., and Secchi, C. (2015).
\newblock An energy tank-based interactive control architecture for autonomous
  and teleoperated robotic surgery.
\newblock \emph{IEEE Transactions on Robotics}, 31(5), 1073--1088.

\bibitem[{Ferraguti et~al.(2019)Ferraguti, Talignani~Landi, Sabattini,
  Bonf{\`e}, Fantuzzi, and Secchi}]{ferraguti2019variable}
Ferraguti, F., Talignani~Landi, C., Sabattini, L., Bonf{\`e}, M., Fantuzzi, C.,
  and Secchi, C. (2019).
\newblock A variable admittance control strategy for stable physical
  human--robot interaction.
\newblock \emph{The International Journal of Robotics Research}, 38(6),
  747--765.

\bibitem[{Ferrari et~al.(2022)Ferrari, Benzi, and Secchi}]{ferrari2022icra}
Ferrari, D., Benzi, F., and Secchi, C. (2022).
\newblock Bidirectional communication control for human-robot collaboration.
\newblock In \emph{2022 International Conference on Robotics and Automation
  (ICRA)}, 7430--7436.
\newblock \doi{10.1109/ICRA46639.2022.9811665}.

\bibitem[{Franken et~al.(2011)Franken, Stramigioli, Misra, Secchi, and
  Macchelli}]{franken2011bilateral}
Franken, M., Stramigioli, S., Misra, S., Secchi, C., and Macchelli, A. (2011).
\newblock Bilateral telemanipulation with time delays: A two-layer approach
  combining passivity and transparency.
\newblock \emph{IEEE transactions on robotics}, 27(4), 741--756.

\bibitem[{Giordano et~al.(2013)Giordano, Franchi, Secchi, and
  Bülthoff}]{giordano2013IJRR}
Giordano, P.R., Franchi, A., Secchi, C., and Bülthoff, H.H. (2013).
\newblock A passivity-based decentralized strategy for generalized connectivity
  maintenance.
\newblock \emph{The International Journal of Robotics Research}, 32(3),
  299--323.
\newblock \doi{10.1177/0278364912469671}.

\bibitem[{Gribovskaya et~al.(2011)Gribovskaya, Kheddar, and
  Billard}]{gribovskaya2011motion}
Gribovskaya, E., Kheddar, A., and Billard, A. (2011).
\newblock Motion learning and adaptive impedance for robot control during
  physical interaction with humans.
\newblock In \emph{2011 IEEE International Conference on Robotics and
  Automation}, 4326--4332. IEEE.

\bibitem[{Hersch et~al.(2008)Hersch, Guenter, Calinon, and
  Billard}]{hersch2008dynamical}
Hersch, M., Guenter, F., Calinon, S., and Billard, A. (2008).
\newblock Dynamical system modulation for robot learning via kinesthetic
  demonstrations.
\newblock \emph{IEEE Transactions on Robotics}, 24(6), 1463--1467.

\bibitem[{LaValle(2006)}]{lavalle_2006}
LaValle, S.M. (2006).
\newblock \emph{Planning Under Differential Constraints}, 587–589.
\newblock Cambridge University Press.
\newblock \doi{10.1017/CBO9780511546877.016}.

\bibitem[{Lecours et~al.(2012)Lecours, Mayer-St-Onge, and
  Gosselin}]{lecours2012variable}
Lecours, A., Mayer-St-Onge, B., and Gosselin, C. (2012).
\newblock Variable admittance control of a four-degree-of-freedom intelligent
  assist device.
\newblock In \emph{2012 IEEE international conference on robotics and
  automation}, 3903--3908. IEEE.

\bibitem[{Lin et~al.(2017)Lin, Liu, Fan, and Tomizuka}]{lin2017real}
Lin, H.C., Liu, C., Fan, Y., and Tomizuka, M. (2017).
\newblock Real-time collision avoidance algorithm on industrial manipulators.
\newblock In \emph{2017 IEEE Conference on Control Technology and Applications
  (CCTA)}, 1294--1299. IEEE.

\bibitem[{Notomista and Egerstedt(2019)}]{notomista2019constraint}
Notomista, G. and Egerstedt, M. (2019).
\newblock Constraint-driven coordinated control of multi-robot systems.
\newblock In \emph{2019 American Control Conference (ACC)}, 1990--1996. IEEE.

\bibitem[{Notomista and Egerstedt(2020)}]{notomista2020persistification}
Notomista, G. and Egerstedt, M. (2020).
\newblock Persistification of robotic tasks.
\newblock \emph{IEEE Transactions on Control Systems Technology}.

\bibitem[{Notomista et~al.(2020)Notomista, Mayya, Selvaggio, Santos, and
  Secchi}]{notomista2020set}
Notomista, G., Mayya, S., Selvaggio, M., Santos, M., and Secchi, C. (2020).
\newblock A set-theoretic approach to multi-task execution and prioritization.
\newblock In \emph{2020 IEEE International Conference on Robotics and
  Automation (ICRA)}, 9873--9879. IEEE.

\bibitem[{Nozaki and Murakami(2009)}]{nozaki2009motion}
Nozaki, K. and Murakami, T. (2009).
\newblock A motion control of two-wheels driven mobile manipulator for
  human-robot cooperative transportation.
\newblock In \emph{2009 35th Annual Conference of IEEE Industrial Electronics},
  1574--1579. IEEE.

\bibitem[{Riggio et~al.(2018)Riggio, Fantuzzi, and Secchi}]{riggio2018use}
Riggio, G., Fantuzzi, C., and Secchi, C. (2018).
\newblock On the use of energy tanks for multi-robot interconnection.
\newblock In \emph{2018 IEEE/RSJ International Conference on Intelligent Robots
  and Systems (IROS)}, 3738--3743. IEEE.

\bibitem[{Secchi and Ferraguti(2019)}]{secchi2019energy}
Secchi, C. and Ferraguti, F. (2019).
\newblock Energy optimization for a robust and flexible interaction control.
\newblock In \emph{2019 International Conference on Robotics and Automation
  (ICRA)}, 1919--1925. IEEE.

\bibitem[{Secchi et~al.(2007)Secchi, Stramigioli, and
  Fantuzzi}]{secchi2007control}
Secchi, C., Stramigioli, S., and Fantuzzi, C. (2007).
\newblock \emph{Control of interactive robotic interfaces: A port-Hamiltonian
  approach}, volume~29.
\newblock Springer Science \& Business Media.

\bibitem[{Siciliano and Villani(2012)}]{siciliano2012robot}
Siciliano, B. and Villani, L. (2012).
\newblock \emph{Robot force control}, volume 540.
\newblock Springer Science \& Business Media.

\bibitem[{Tagliabue et~al.(2017)Tagliabue, Kamel, Verling, Siegwart, and
  Nieto}]{tagliabue2017collaborative}
Tagliabue, A., Kamel, M., Verling, S., Siegwart, R., and Nieto, J. (2017).
\newblock Collaborative transportation using mavs via passive force control.
\newblock In \emph{2017 IEEE international conference on robotics and
  automation (ICRA)}, 5766--5773. IEEE.

\end{thebibliography}
\end{document}